\documentclass{article}

\usepackage{PRIMEarxiv}

\usepackage[utf8]{inputenc} 
\usepackage[T1]{fontenc}    
\usepackage{hyperref}       
\usepackage{url}            
\usepackage{booktabs}       
\usepackage{amsfonts}       
\usepackage{nicefrac}       
\usepackage{microtype}      
\usepackage{lipsum}
\usepackage{fancyhdr}       
\usepackage{graphicx}       
\graphicspath{{media/}}     

\usepackage{algorithm}
\usepackage{amsmath}
\usepackage{amssymb}
\usepackage{multirow}
\usepackage{algcompatible}
\usepackage{natbib}
\usepackage{hyperref}

\usepackage{mdframed}
\usepackage{listings}
\usepackage{xcolor}
\newtoggle{showblue}
\togglefalse{showblue}  

\newcommand{\maybeBlue}[1]{%
  \iftoggle{showblue}{\textcolor{blue}{#1}}{#1}%
}

\makeatletter
\newcommand{\equalcontrib}{\textsuperscript{\dag}}
\newcommand{\corresponding}{\textsuperscript{\ddag}}
\makeatother

\title{CAMA: Enhancing Mathematical Reasoning in Large Language Models with Causal Knowledge
\\
{\small Version Including Appendices}
}

\author{
  Lei Zan\equalcontrib\corresponding   \\
  Huawei Noah's Ark Lab \\
  France\\
  \texttt{xdzanlei@gmail.com} \\
  \And
  Keli Zhang\equalcontrib\corresponding  \\
  Huawei Noah's Ark Lab \\
  France\\
  \texttt{zhangkeli1@huawei.com} \\
  \And
  Ruichu Cai \\
  Guangdong University of Technology \\
  Peng Cheng Laboratory \\
  China \\
  \texttt{cairuichu@gmail.com} \\
  \And
  Lujia Pan \\
  Huawei Noah's Ark Lab \\
  China\\
  \texttt{panlujia@huawei.com} \\
}

\begin{document}
\maketitle

\begingroup
\renewcommand{\thefootnote}{\fnsymbol{footnote}}
\footnotetext[2]{Equal contribution.}
\footnotetext[3]{Corresponding author.}
\footnotetext[4]{This paper was accepted at AAAI 2026 (Main Track).}
\endgroup

\begin{abstract}
  Large Language Models (LLMs) have demonstrated strong performance across a wide range of tasks, yet they still struggle with complex mathematical reasoning, a challenge fundamentally rooted in deep structural dependencies. To address this challenge, we propose \textbf{CA}usal \textbf{MA}thematician (\textbf{CAMA}), a two-stage causal framework that equips LLMs with explicit, reusable mathematical structure. \maybeBlue{In the learning stage, CAMA first constructs the \textbf{M}athematical \textbf{C}ausal \textbf{G}raph (\textbf{MCG}), a high-level representation of solution strategies, by combining LLM priors with causal discovery algorithms applied to a corpus of question-solution pairs. The resulting MCG encodes essential knowledge points and their causal dependencies. To better align the graph with downstream reasoning tasks, CAMA further refines the MCG through iterative feedback derived from a selected subset of the question-solution pairs.} In the reasoning stage, given a new question, CAMA dynamically extracts a task-relevant subgraph from the MCG, conditioned on both the question content and the LLM’s intermediate reasoning trace. This subgraph, which encodes the most pertinent knowledge points and their causal dependencies, is then injected back into the LLM to guide its reasoning process. Empirical results on real-world datasets show that CAMA significantly improves LLM performance on challenging mathematical problems. Furthermore, our experiments demonstrate that structured guidance consistently outperforms unstructured alternatives, and that incorporating asymmetric causal relationships yields greater improvements than using symmetric associations alone.
\end{abstract}

\keywords{Large language model \and Causal graph model \and Mathmatical reasoning}

\begin{itemize}
    \item \textbf{Code:} \href{https://github.com/huawei-noah/trustworthyAI/tree/master/research/CAMA}{https://github.com/huawei-noah/trustworthyAI/tree/master/research/CAMA}
\end{itemize}

Large Language Models (LLMs), such as the GPT series~\citep{openai2024gpt4technicalreport}, DeepSeek series~\citep{deepseekai2025deepseekv3technicalreport, deepseekai2025deepseekr1incentivizingreasoningcapability}, and Pangu series~\citep{yin2025panguultrapushinglimits, tang2025panguultramoetrain}, have achieved remarkable advances across a wide spectrum of language tasks, including question answering, code synthesis, and information retrieval~\citep{ouyang2022training, gu2023llm, dai2024bias}. 

Despite these advances, LLMs still struggle with challenging mathematical problems~\citep{han2023veritymath}, where solutions require formal rigor, symbolic manipulation, and multi‑step deduction~\citep{cobbe2021training}. This limitation can be attributed primarily to two key challenges. First, the inherent architectural constraints of transformers impose a fixed depth on reasoning, limiting the model’s ability to carry out deep and interdependent logical inferences~\citep{merrill2023parallelism, liu2024much}. Second, LLMs primarily rely on statistical pattern recognition, making them sensitive to minor changes in problem phrasing and prone to brittle or inconsistent outputs~\citep{mirzadeh2024gsm, jiang2024peek}.

To address these challenges, we advocate for a shift from purely data-driven prediction toward structured reasoning guidance. Drawing inspiration from the dictum “teach a man to fish rather than give him a fish,” we argue that equipping a model with explicit, reusable problem‑solving strategies is an efficient way to strengthen its reasoning abilities. To this end, we formalize such strategies as a \textbf{M}athematical \textbf{C}ausal \textbf{G}raph (\textbf{MCG}): a directed acyclic graph in which each node is a knowledge point (\textit{e.g.}, \textit{Area of a circle}) and each edge encodes a causal dependency (\textit{e.g.}, computing \textit{Volume of a cylinder} first requires the circle’s area). \maybeBlue{The MCG stores information that is generally applicable across contexts, and the edge direction specifies the order of reasoning, which is especially valuable when the required chain of knowledge spans many steps. Integrating such a graph into the prompt helps decompose complex problems into coherent subtasks, reduce reliance on implicit reasoning, and increase intermediate accuracy, while curbing hallucinations and redundancy.}

Building on these insights, we present \textbf{CA}usal \textbf{MA}thematician (\textbf{CAMA}), a plug-and-play framework that integrates causal discovery with LLMs to construct and exploit MCGs. CAMA operates in two stages: (1) in the learning stage, CAMA parses LLM‑generated chain-of-thought solutions to extract underlying knowledge points, then infers the causal dependencies among them. The graph is subsequently refined via feedback from the model’s own question‑answering accuracy. (2) In the reasoning stage, CAMA feeds this structured information back into the LLM via natural-language prompts to boost problem-solving performance, without the need for any parameter updates, such as supervised fine-tuning, making it lightweight and adaptable. 

Our main contributions are as follows:
\begin{itemize}
    \item We propose the \textbf{M}athematical \textbf{C}ausal \textbf{G}raph (\textbf{MCG}), a reusable, high-level representation that captures the causal structure of mathematical solution strategies.
    \item We develop \textbf{CAMA}, a lightweight and plug‑and‑play framework that combines LLMs with causal discovery to automatically derive and utilize MCGs for reasoning-enhanced problem solving.
    \item We empirically demonstrate that CAMA outperforms standard prompting methods on real-world datasets. In particular, structured, directed guidance via MCGs leads to more reliable and accurate solutions than unstructured or symmetric alternatives.
\end{itemize}

\begin{figure*}[!h]
  \centering
  \includegraphics[width=\textwidth, trim=0.5cm 2.8cm 4cm 0.4cm, clip]{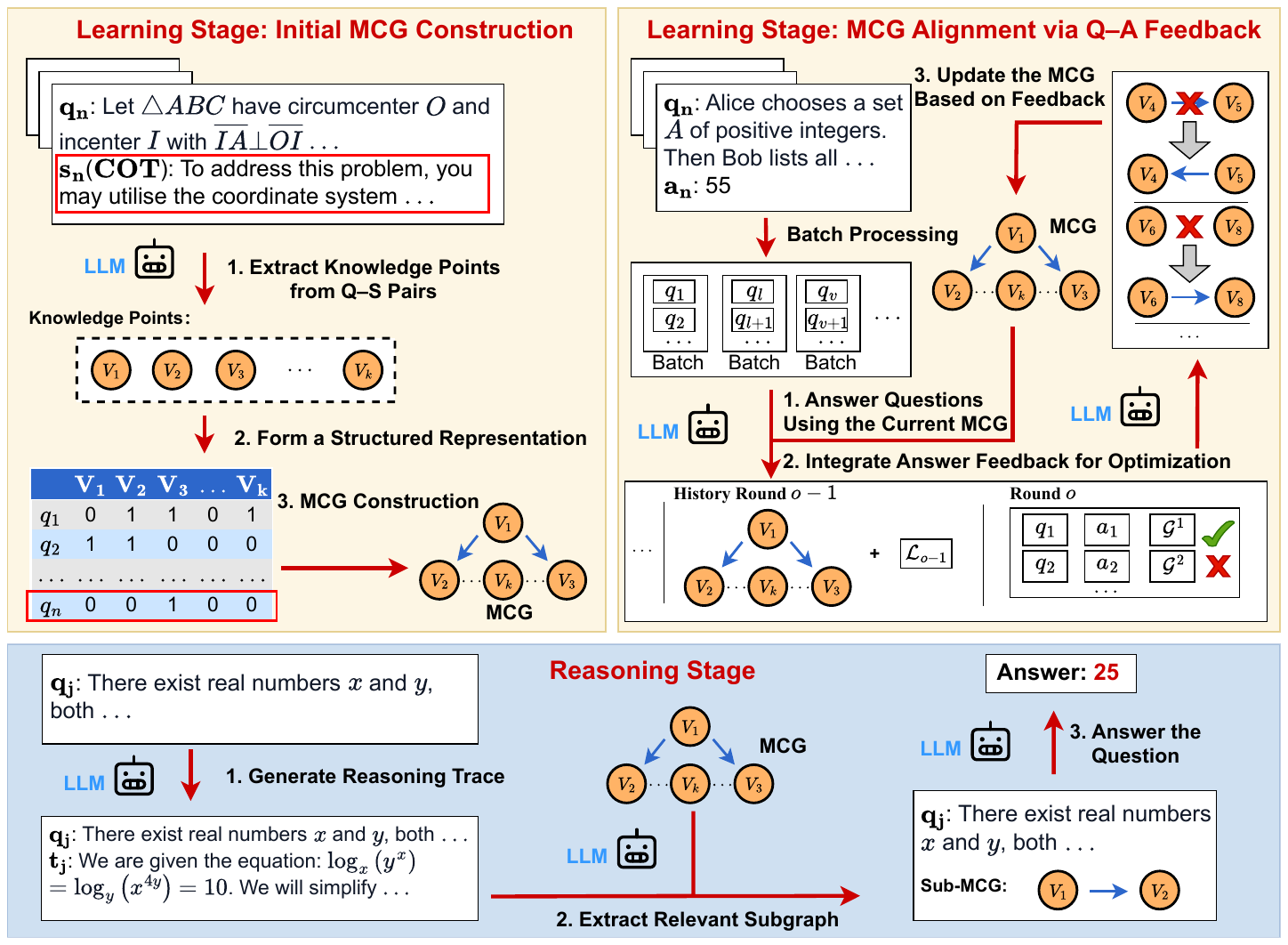}
  \caption{The CAMA framework consists of two stages: learning and reasoning. In the learning stage, (1) CAMA constructs an initial Mathematical Causal Graph (MCG) from question–solution pairs by combining LLM outputs with classical causal discovery methods to identify key knowledge points and their causal dependencies; (2) the MCG is then refined using feedback from the LLM’s answers to better align with the downstream reasoning task. In the reasoning stage, the optimized MCG is used to solve new questions through a three-step process: generating a reasoning trace, extracting a relevant subgraph, and guiding the LLM to produce the final answer.
}
  \label{fig:overview_cama}
\end{figure*}
\section{Related work}

\subsection{Causal discovery} 

Identifying causal relationships is a fundamental problem across many empirical sciences. Traditionally, such relationships are inferred through interventions or randomized controlled experiments. However, these approaches are often costly, time-consuming, or even infeasible in practice. As an alternative, causal discovery algorithms aim to infer a causal graph, such as a directed acyclic graph (DAG), from passively collected observational data that are typically easier to obtain. A wide range of causal discovery methods have been developed, falling into several main categories: constraint-based methods (\textit{e.g.}, PC, FCI)~\citep{spirtes2000causation, spirtes1995fci}, which rely on conditional independence tests; noise-based methods (\textit{e.g.}, LiNGAM)~\citep{shimizu2011directlingam, peters2017elements}, which exploit asymmetric patterns of noise in the data; score-based methods (\textit{e.g.}, GES)~\citep{chickering2002optimal}, which search for high-scoring graph structures; and optimization-based methods (\textit{e.g.}, NOTEARS)~\citep{zheng2018dags}, which frame DAG learning as a continuous optimization problem. For a comprehensive review of these methods, please refer to these surveys~\citep{assaad2022survey, glymour2019review}. In addition, several practical Python libraries have emerged to support causal discovery~\citep{zhang2021gcastlepythontoolboxcausal, zheng2024causal}. Despite their theoretical appeal, practitioners should remain mindful that each method’s assumptions may not always hold in complex environments~\citep{ait2023case}. For example, the PC algorithm~\citep{spirtes2000causation} assumes \textit{faithfulness} and \textit{causal sufficiency}~\citep{spirtes2000causation}, meaning that all statistical independencies within the observed data are encoded in the causal graph and there are no unobserved hidden common causes. Moreover, some methods, such as PC and GES, typically recover only a Markov equivalence class (MEC) of DAGs, which includes all graphs that encode the same set of conditional independencies. These equivalence classes are commonly represented by a completed partially directed acyclic graph (CPDAG)~\citep{chickering2002learning}. 

Recently, the rapid advancement of LLMs and the availability of vast amounts of unstructured data, particularly text, have drawn significant attention to leveraging LLMs to extract causal concepts and identify the relationships among them~\citep{scholkopf2021toward, causalcoat2024, wang2025causal}.

\subsection{Reinforcing mathematical reasoning ability of LLMs}
Improving the mathematical reasoning ability of LLMs has recently attracted significant attention, with advancements emerging from both pre-training and post-training strategies. Math-specific pre-trained models such as Llemma~\citep{azerbayev2023llemma}, DeepSeekMath~\citep{shao2024deepseekmath}, InternLM-Math~\citep{ying2024internlm}, and Qwen2-Math~\citep{yang2024qwen2} enhance performance by training on curated, math-rich corpora. These models benefit from data sourced from scientific texts, programming content, and formal mathematical proofs, enabling them to develop a deeper domain understanding. In addition, post-training techniques further refine reasoning through supervised fine-tuning on specialized datasets. Program-of-Thought (PoT)~\citep{chen2022program}, evol-Instruct~\citep{luo2023wizardmath}, and Tool-Integrated Reasoning (TIR)~\citep{gou2023tora} are representative approaches that teach models to solve problems step-by-step, often by generating executable code or integrating external tools such as Python for reliable calculations. Models such as WizardMath~\citep{luo2023wizardmath} and MetaMath~\citep{yu2023metamath} exemplify how instruction tuning and reasoning format alignment significantly improve math benchmark performance. Beyond supervised learning, preference-based methods like Step-DPO~\citep{lai2024step} and online RLHF~\citep{wang2025causal} focus on optimizing the reasoning process itself by learning from step-level feedback. These strategies train models to prefer correct intermediate steps, rather than only focusing on final answers, resulting in more robust and interpretable solutions. While these efforts have led to substantial improvements, they primarily rely on statistical learning or preference signals. 

\section{The CAMA Framework}
\label{sec:metho}
\subsection{Problem Setup}
\paragraph{Dataset Description} We are given a dataset $\mathbf{D}=\{(q_i,s_i,a_i)\}_{i=1}^n$ of $n$ independently drawn triples from the product space $\mathbf{Q}\times \mathbf{S}\times \mathbf{A}$, where  $q_i\in\mathbf{Q}$ is an unstructured question, $s_i\in\mathbf{S}$ is the corresponding detailed solution, and $a_i\in\mathbf{A}$ is the ground‑truth answer (integer or symbolic).

\paragraph{Learning Objective} The goal of CAMA is to learn a mapping 
$$
f : (\mathbf Q \times \mathbf S \times \mathbf A)^{n} \to \mathbf G,\qquad f(\{(q_i,s_i, a_i)\}_{i=1}^n)=\mathcal G.
$$
However, most existing mathematical datasets do not provide detailed solution steps. To address this limitation, we leverage an LLM to generate a chain-of-thought solution $s_i$ for each question $q_i$, thereby constructing the solution set $\{s_i\}_{i=1}^n$. 

We denote the Mathematical Causal Graph (MCG) as $\mathcal{G} = (\mathbf{V}, \mathbf{E}) \in \mathbf{G}$, with its components described in detail in the following section. In this graph, $\mathbf{V} = \{V_1, \dots, V_k\}$ is the set of $k$ nodes, each corresponding to a distinct mathematical knowledge point (\textit{e.g.}, \textit{Volume of a cylinder}, \textit{Area of a circle}), and $\mathbf{E}$ is the set of edges capturing the causal relationships among these knowledge points. 

To construct $\mathcal{G}$, we first learn an intermediate mapping
$$
h : (\mathbf Q \times \mathbf S)^{n} \to \{0,1\}^{n \times k},\qquad h(\{(q_i,s_i)\}_{i=1}^n)=\mathbf Z,
$$ 
where $\mathbf{Z} \in \{0,1\}^{n \times k}$. Each row corresponds to a question–solution pair, and each column to a knowledge point. The element $Z_{i,j} \in \{0, 1\}$ indicates whether knowledge point $V_j$ is required to solve question $q_i$: $Z_{i,j} = 1$ if the knowledge point $V_j$ is necessary for solving question $q_i$; otherwise $Z_{i,j} = 0$. In practice, $\mathbf Z$ is obtained through a multi-step pipeline consisting of knowledge point extraction, deduplication, and parsing (via the function $l(\cdot)$ defined later). Thus, the mapping $h(\cdot)$ should be regarded as a composite procedure rather than a separate model. To extract and formalize the set of knowledge points $\mathbf{V} = \{V_1, \dots, V_k\}$ from the input pairs $\{(q_i, s_i)\}_{i=1}^{n}$, we leverage the language understanding and domain knowledge capabilities of the LLM. Once the binary matrix $\mathbf{Z}$ is obtained, a causal discovery algorithm (denoted by $\mathrm{CD}$) is applied to infer the structure of the graph. This graph is then iteratively refined using feedback from a selected subset of question–answer pairs, allowing it to better align with downstream reasoning tasks.

\begin{figure}[t]
    \centering
    \includegraphics[width=0.5\columnwidth]{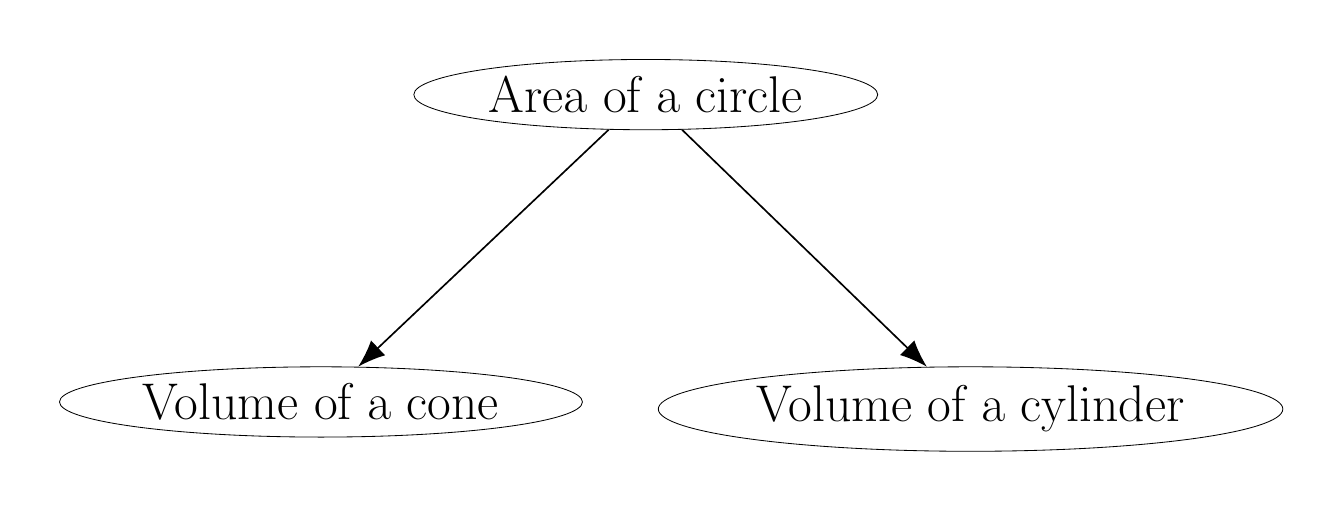} 
    \caption{An example of a Mathematical Causal Graph (MCG) is shown, illustrating three knowledge points: \textit{Area of a circle}, \textit{Volume of a cylinder}, and \textit{Volume of a cone}. The edges indicate that understanding the \textit{Area of a circle} is required to compute both the \textit{Volume of a cylinder} and the \textit{Volume of a cone}.
    } %
    \label{fig:MCG}
\end{figure}

\paragraph{Mathematical Causal Graph (MCG)} We represent high-level solution strategies for mathematical problems using a Mathematical Causal Graph (MCG), denoted as $\mathcal{G} = (\mathbf{V}, \mathbf{E})$. In this graph, nodes correspond to relevant theorems or definitions, while edges capture causal relationships and the potential order in which these knowledge points are applied during problem solving. For simplicity and without loss of generality, we assume that the MCG is a directed acyclic graph. Figure~\ref{fig:MCG} illustrates an example MCG with three knowledge points: \textit{Area of a circle}, \textit{Volume of a cylinder}, and \textit{Volume of a cone}. A directed edge from \textit{Area of a circle} to \textit{Volume of a cylinder} (and similarly to \textit{Volume of a cone}) indicates that computing the area of the circle is a prerequisite for determining the cylinder's volume. More generally, each edge represents a possible (though not necessarily unique) reasoning path. 

\paragraph{Mathematical Dataset Construction}
To construct the Mathematical Causal Graph (MCG), it is crucial to curate the dataset $\mathbf{D}=\{(q_i, s_i, a_i)\}_{i=1}^{n}$. We begin with a readily available dataset $\mathbf{QA} = \{(q_i, a_i)\}_{i=1}^{n}$. For each question $q_i$, we employ an LLM (denoted $\mathcal{LLM}$) together with a prompt template $p_g$ (see Appendix~\ref{app:used_prompts} for details on this and other templates) to generate a detailed chain-of-thought solution $s_i$ and a predicted answer $\hat{a}_i$: 
$$
s_i, \hat{a}_i = \mathcal{LLM}(q_i, p_g).
$$ 
The solution $s_i$ is retained and paired with $q_i$ only if the predicted answer matches the ground truth, \textit{i.e.}, $\hat{a}_i = a_i$. By concatenating each retained solution with its corresponding question, we obtain the set of question–solution pairs $\mathbf{QS} = \{(q_i, s_i)\}_{i=1}^{n}$.

\subsection{Overview of the CAMA Framework}
This section introduces the two-stage CAMA framework, illustrated in Figure~\ref{fig:overview_cama}, comprising a learning stage and a reasoning stage. In the learning stage, CAMA first builds an initial Mathematical Causal Graph (MCG) from a set of question–solution pairs and \maybeBlue{then refines it using feedback from a downstream reasoning task.} In the reasoning stage, a new question is solved in three sequential steps guided by the MCG, ultimately producing the final answer.

\subsection{Learning Stage}
The learning stage of the CAMA framework consists of two key components: (1) constructing an initial Mathematical Causal Graph (MCG) from question–solution pairs, and (2) refining the graph by aligning it with the reasoning task through LLM-based feedback. The corresponding pseudocode is provided in Appendix~\ref{app:code_learning}. 

\subsubsection{Initial MCG Construction} This step involves three main sub-tasks: extracting knowledge points from question–solution pairs, forming a structured representation, and applying causal discovery to infer the initial MCG. The input consists of $n$ pairs: $\mathbf{QS} = \{(q_i, s_i)\}_{i=1}^{n}$. For each pair $(q_i, s_i)$, $\mathcal{LLM}$ is prompted with $p_p$ to extract up to $\lambda$ relevant knowledge points: 
$$
\mathbf{V}^i = \mathcal{LLM}(q_i, s_i, \lambda, p_p).
$$ 
Each knowledge point $V_j^i \in \mathbf{V}^i$ is a tuple: $V_j^i = (V_{j, key}^i,V_{j, des}^i),$
where $V_{j, key}^i$ is a short label, and $V_{j, des}^i$ is a detailed description. The parameter $\lambda$ controls the granularity of extracted knowledge points: smaller $\lambda$ leads the LLM to produce broader, more general knowledge points, while larger $\lambda$ allows for more specific and detailed extractions tailored to each question–solution pair. Given the sensitivity of LLMs to prompt phrasing and input order~\citep{razavi2025promptsensitivity, mirzadeh2025gsm}, knowledge extraction is performed independently for each question to avoid prompt interference. This results in a set of extracted knowledge points for each item in the dataset. The union of all extracted sets is: 
$$
\mathbf{V}' = \bigcup_{i=1}^{n} \mathbf{V}^i.
$$ 
The aggregated set $\mathbf{V}'$ may contain redundant or semantically overlapping entries. To address this, we employ another $\mathcal{LLM}$ with prompt template $p_r$ to identify and remove redundant knowledge points: 
$$
\mathbf{V}, \mathbf{O} = \mathcal{LLM}(\mathbf{V}', p_r),
$$ 
where $\mathbf{V}$ is the deduplicated set, and each pair $(V_i, V_j) \in \mathbf{O}$ indicates that $V_j$, an element in $\mathbf{V}' \setminus \mathbf{V}$, can be replaced by $V_i \in \mathbf{V}$. Finally, we reconstruct the structured representation $\mathbf{Z}$ by parsing the original extracted sets $\{\mathbf{V}^1, \dots, \mathbf{V}^n\}$ with $\mathbf{V}$ using the replacement mapping $\mathbf{O}$. Specifically, each row of $\mathbf{Z}$ corresponds to a question–solution pair $(q_i, s_i)$, and each column represents a knowledge point in $\mathbf{V}$. A column is set to 1 if the corresponding knowledge point either appears in $\mathbf{V}^i$ or replaces an element of $\mathbf{V}^i$ according to $\mathbf{O}$; otherwise, it is set to 0. This parsing process is defined by the function $l(\cdot)$: 
$$
\mathbf{Z} = l(\{\mathbf{V}^1, \dots, \mathbf{V}^n\}, \mathbf{V}, \mathbf{O}).
$$
After obtaining the binary matrix $\mathbf{Z}$ from the question–solution pairs $\mathbf{QS}$, we apply a causal discovery method ($\mathrm{CD}$), to infer the MCG: 
$$
\mathcal{G} = \mathrm{CD}(\mathbf{Z}).
$$ 
In our experiments, we employ the PC algorithm, which yields a completed partially directed acyclic graph (CPDAG) comprising both directed and undirected edges. The resulting edges should be viewed as dataset-level averages, links appearing in only a few instances are generally pruned. This method relies on two standard assumptions: \textit{causal sufficiency} — no unobserved knowledge point acts as a common prerequisite for any pair in $\mathbf{V}$, and \textit{faithfulness} — the inferred graph $\mathcal{G}$ captures all dependencies among the knowledge points. Although we adopt PC here for concreteness, our framework is agnostic to the specific causal discovery method and can accommodate alternatives such as FCI, provided that their assumptions align with the data.

\subsubsection{MCG Alignment via Question–Answer Feedback} To adapt the edge set $\mathbf{E}$ of the MCG $\mathcal{G}$ to downstream reasoning tasks, we iteratively update $\mathcal{G}$ with feedback from $\mathcal{LLM}$, which generates answers conditioned on the current graph. The alignment aims to maximize $\mathcal{LLM}$’s precision on a designated subset of question–answer pairs. \maybeBlue{Formally we seek
$$
\mathcal{G}^* = \arg\max_{\mathcal{G}} \; \mathbb{E}_{(q_i, s_i, a_i) \sim \mathbf{D}^s} \left[ \mathbf{1}_{\{\mathcal{LLM}(\mathcal{G}, q_i, p_a) = a_i\}} \right],
$$
where $\mathbf{D}^{s}=\{(q_i,s_i, a_i)\}_{i=1}^{m}$ denotes a subset of $m$ samples drawn from $\mathbf{D}$, for each question $q_i$ the model uses $\mathcal{G}$ through the prompt template $p_a$ and returns 
$$
\hat a_i=\mathcal{LLM}(\mathcal{G},q_i,p_a),
$$ 
the indicator function $\mathbf{1}_{\{\cdot\}}$ outputs 1 when $\hat a_i$ matches the ground truth $a_i$ and 0 otherwise.}

To solve this problem, we adopt a batch-based iterative optimization procedure over $n_e$ epochs. At each optimization round $o$, we draw a batch of $s_b$ samples from $\mathbf{D}^{s}$:
$$
\mathbf{D}'_{o} = \{(q_j, s_j, a_j)\}_{j=1}^{s_b} \subset \mathbf{D}^{s}.
$$
Instead of using the full graph $\mathcal{G}_o$, we extract a subgraph $\mathcal{G}^{j}_{o} \subseteq \mathcal{G}_o$ for each question $q_j$ that includes only relevant knowledge points. To do this, we first ask the $\mathcal{LLM}$ using prompt template $p_t$ to analyze the question and produce a candidate chain-of-thought solution:
$$
t_j = \mathcal{LLM}(q_j, p_t).
$$ 
We then use the generated reasoning trace $t_j$ along with $q_j$ and the full graph $\mathcal{G}_o$ to generate the matched subgraph using prompt $p_m$: 
$$
\mathcal{G}^{j}_{o} = \mathcal{LLM}(\mathcal{G}_o, t_j, q_j, p_m).
$$
The model is then queried as follows:
$$
\hat{a}_j = \mathcal{LLM}(\mathcal{G}^{j}_{o}, q_j, p_a).
$$
We record the quadruple $(q_j,s_j,\mathcal{G}^{j}_{o},\mathbf{1}_{\{\hat a_j=a_j\}})$ and repeat the process for all $s_{b}$ questions. 

\maybeBlue{After the batch is finished, these quadruples together with an optimization history over $r$ previous rounds:
$$
\mathbf{H}=\{(\mathcal{G}_{o-1},\mathcal{L}_{o-1}),\dots,(\mathcal{G}_{o-r},\mathcal{L}_{o-r})\}
$$ 
are given to the model through the update prompt $p_{u}$ to obtain the next graph:
$$
\mathcal{G}_{o+1} = \mathcal{LLM}(\{(q_j, s_j, \mathcal{G}^j_{o}, \mathbf{1}_{\{\hat{a}_j = a_j\}})\}_{j=i}^{s_b} , \mathbf{H}, p_u),
$$
Edges that support correct reasoning are reinforced, whereas those linked to errors are revised. Here, $\mathcal{L}_{o-1}$ represents the precision achieved in the previous round with graph $\mathcal{G}_{o-1}$:
$$
\mathcal{L}_{o-1} = \frac{1}{s_b} \sum_{j=1}^{s_b} \mathbf{1}_{\{\hat{a}_j = a_j\}}.
$$ }

At the end of each epoch, we evaluate the precision of $\mathcal{G}_{o+1}$ over $\mathbf{D}^s$. After all epochs, the graph with the highest precision on $\mathbf{D}^s$ is selected as the final optimized graph $\mathcal{G}^*$. To improve efficiency, the optimization process is terminated early if the graph remains unchanged for $c_{stop}$ consecutive batches. We further assume that MCGs derived from the alignment and downstream samples share partial structural overlap, enabling strategy preferences learned during alignment to transfer effectively to downstream reasoning.

\subsection{Reasoning Stage}
In the reasoning stage, a new question is processed in three successive steps using the MCG $\mathcal{G}^*$, ultimately yielding the final answer. This procedure mirrors the alignment process used during training. The pseudocode is presented in Appendix~\ref{app:code_reasoning}.

\paragraph{Generate Reasoning Trace}
Given a new question $q_j$, we first prompt $\mathcal{LLM}$ using a template $p_t$ to produce a candidate chain-of-thought solution:
$$
t_j = \mathcal{LLM}(q_j, p_t).
$$  
\paragraph{Extract Relevant Subgraph}
Next, the generated reasoning trace $t_j$ is used, along with the question $q_j$ and the full graph $\mathcal{G}^*$, to extract a relevant subgraph via the prompt template $p_m$:
$$
\mathcal{G}^{j} = \mathcal{LLM}(\mathcal{G}^*, t_j, q_j, p_m).
$$
\paragraph{Answer the Question}
Finally, the model is prompted with the matched subgraph $\mathcal{G}^j$, the question $q_j$, and the answering prompt $p_a$ to generate the predicted answer:
$$
\hat{a}_j = \mathcal{LLM}(\mathcal{G}^{j}, q_j, p_a).
$$
To enable effective reasoning, in practice, the subgraph $\mathcal{G}^j$ is encoded into natural language. Each edge between knowledge points is verbalized. For instance, a directed edge from \textit{Area of a circle} to \textit{Volume of a cylinder} is expressed as: "\textit{Area of a circle} is a prerequisite for \textit{Volume of a cylinder}. If \textit{Volume of a cylinder} is used, then \textit{Area of a circle} could also be used." If the edge is undirected, we phrase it as: "\textit{Area of a circle} and \textit{Volume of a cylinder} are associated, but the direction of dependency is unclear. Either could be a prerequisite for the other." This verbalization follows directly from the way we build the structured matrix $\mathbf{Z}$. For each question–solution pair, a knowledge point’s presence is encoded as 1 and its absence as 0. A directed edge from one knowledge point to another therefore means the source point always appears whenever the target does, allowing us to interpret the source as a prerequisite for the target.

\section{Experiments}
To assess the effectiveness of CAMA in enhancing the LLM's mathematical reasoning, we conduct the following experiments. 

\paragraph{Datasets}
We evaluate our framework on three mathematical benchmarks: AIME, Omni-MATH~\cite{gao2024omni}, and OlympiadBench~\cite{he2024olympiadbench}. The American Invitational Mathematics Examination (AIME) is a high-level competition for high school students, with 30 questions per year spanning four categories: algebra, geometry, number theory, and combinatorics. Omni-MATH is a recent Olympiad-style benchmark containing 4,428 problems across nine subdomains. For our experiments, we use a 200-question subset, referred to as Omni-MATH-200, which focuses on the same four categories. OlympiadBench includes 8,476 international Olympiad problems. We evaluate on a filtered subset of 674 English, non-proof questions, called OlympiadBench-674, which is also grouped into the same four categories. The distribution of questions in these datasets is shown in Appendix~\ref{app:dataset_distribution}.

\paragraph{Evaluation Setup and Compared Methods}
We adopt DeepSeek-V3-0324~\citep{deepseekai2025deepseekv3technicalreport} (\textbf{DSV3}) and Qwen3-32B~\citep{yang2025qwen3} (\textbf{Qwen3}) as our base LLMs, using a temperature of 0.6. DSV3 is accessed via API, whereas Qwen3 is deployed locally. The version enhanced with our proposed CAMA framework is denoted \textbf{CAMA}. Unless otherwise specified, we set the knowledge point granularity to $\lambda = 3$, batch size to $s_b = 5$, number of epochs to $n_e = 10$, history length to $r = 7$, and early stopping threshold to $c_{stop} = 3$. We also evaluate two ablated variants: \textbf{CAMA \textit{w/o} Alignment}, which excludes the alignment stage, and \textbf{CAMA \textit{w/o} Directed Edge}, in which all directed edges in the Mathematical Causal Graph (MCG) are replaced with undirected edges. For comparison, we include two baselines based on the chain-of-thought prompting~\citep{wei2022chain}: \textbf{COT-ZeroShot}, which uses no examples, and \textbf{COT-FewShot}, which uses the full AIME2023 dataset along with detailed solutions as in-context examples (due to input token length limitations, AIME2022 questions are excluded). To construct the Mathematical Causal Graph (MCG), we apply the PC algorithm from the \texttt{gCastle} library~\citep{zhang2021gcastlepythontoolboxcausal} with the G-squared test and default settings.

\paragraph{Evaluation Metric}
Model performance is evaluated with the Pass@1 metric, defined as the fraction of test questions answered correctly on the first attempt. A prediction is considered correct if it exactly matches or is mathematically equivalent to the ground truth. Formally, given a test set $\mathbf{Q} = \{q_1, \dots, q_n\}$, let $\hat{a}_i$ be the model’s answer and $a_i$ the ground truth for $q_i$. An indicator function $\text{judge}(\hat{a}_i, a_i)$ returns 1 if the answer is correct, and 0 otherwise, yielding $\text{Pass@1} = \frac{1}{n} \sum_{i=1}^n \text{judge}(\hat{a}_i, a_i).$


\paragraph{Experimental Setup}
To construct the MCG, we use a training set of 60 questions sourced from AIME2022 and AIME2023. The resulting graph learned by \textbf{CAMA} is visualized in Appendix~\ref{app:mcg}. We prompt DeepSeek-R1~\citep{deepseekai2025deepseekr1incentivizingreasoningcapability} to produce detailed chain-of-thought solutions for each question. Alignment is performed using the same data as the training set.
Model performance is then evaluated on four test sets: AIME2024, AIME2025, Omni-MATH-200, and OlympiadBench-674. 
Given that answers in the AIME datasets are purely numeric, correctness is evaluated using a strict equality check: $\text{judge}(\hat{a}_i, a_i) = 1$ if and only if $\hat{a}_i = a_i$. For Omni-MATH-200 and OlympiadBench-674, where answers may include symbolic expressions, we employ Omni-Judge~\citep{gao2024omni}, an evaluator built on LLaMA-3.1-8B-Instruct, to assess correctness.

\begin{table*}[!h]
\centering
\begin{tabular}{c|c|c|c|c|c}
\toprule
\multirow{2}{*}{Base LLM} & \multirow{2}{*}{Method} &  AIME2024 & AIME2025 & Omni-MATH-200 & OlympiadBench-674 \\ 
&  & Pass@1($\uparrow$) & Pass@1($\uparrow$) & Pass@1($\uparrow$) & Pass@1($\uparrow$)  \\ \midrule  
\multirow{6}{*}{\textbf{DSV3}} & \textbf{COT-ZeroShot} & 42.2\% & 35.6\% & 42.2\% & \textbf{67.2\%} \\
& \textbf{COT-FewShot} & 43.3\% & 31.0\% & 41.0\% & 65.2\% \\
& \textbf{DSV3} & 39.2\% & 28.8\% & 42.0\% & 65.0\% \\
& \textbf{CAMA \textit{w/o} Directed Edge} & 43.3\% & 33.3\%  & 43.7\% & 65.8\% \\
& \textbf{CAMA \textit{w/o} Alignment} & 47.8\% & \textbf{38.9\%} & 43.0\% & 66.8\% \\
& \textbf{CAMA} (ours) & \textbf{50.0\%} &  \textbf{38.9\%} & \textbf{45.0\%} & 66.4\%   \\ 
\midrule
\multirow{6}{*}{\textbf{Qwen3}} & \textbf{COT-ZeroShot} & 75.6\% & 72.2\% & 68.7\% & 81.4\% \\
& \textbf{COT-FewShot} & 66.7\% & 61.1\% & 58.5\% & 77.1\%\\
& \textbf{Qwen3} & 74.4\% & 65.6\% & 64.2\% & 81.2\%\\
& \textbf{CAMA \textit{w/o} Directed Edge} & 75.6\% & 72.2\% & 67.0\% & 82.0\% \\
& \textbf{CAMA \textit{w/o} Alignment} & 78.9\% & 74.4\% & 67.3\% & 82.2\% \\
& \textbf{CAMA} (ours) & \textbf{82.2\%} & \textbf{76.7\%} & \textbf{69.0\%} & \textbf{83.3\%} \\
\bottomrule
\end{tabular}

\caption{This table reports Pass@1 scores for each method on four datasets using DeepSeek-V3-0324 (\textbf{DSV3}) and Qwen3-32B (\textbf{Qwen3}). Each value is averaged over three repetitions. Omni-MATH-200 is a 200-question subset of Omni-MATH covering four categories. Likewise, OlympiadBench-674 denotes a 674-question English, non-proof subset of OlympiadBench. Bold values indicate better performance.}
\label{tab:pass1}
\end{table*}

\paragraph{Results}  
Table~\ref{tab:pass1} presents the Pass@1 scores of each method on four benchmark datasets using two base LLMs, with each result averaged over three repetitions (the corresponding standard deviations are provided in Appendix~\ref{app:std_passe_1_score}). Overall, \textbf{CAMA} leads in most cases. The comparisons across method variants reveal several key observations. First, \textbf{CAMA} consistently outperforms \textbf{COT-FewShot}, indicating that structured information encoded in the Mathematical Causal Graph (MCG) is more effective for enhancing mathematical reasoning than raw text prompting. Second, the comparison between \textbf{CAMA} and \textbf{CAMA w/o Directed Edge} demonstrates the importance of asymmetric (directed) relations in the MCG over purely symmetric ones, confirming the value of capturing causal directionality. Third, the comparison with \textbf{CAMA w/o Alignment} underscores the benefit of the alignment step in refining the graph and adapting it more closely to the LLM's reasoning needs. Finally, the relatively moderate performance of \textbf{CAMA} based on \textbf{DSV3} on OlympiadBench-674 can be partially attributed to limited utilization of the benchmark's knowledge components in the MCG. This limitation can be eased by adopting a coarser knowledge point granularity: with $\lambda = 2$, \textbf{CAMA} attains a Pass@1 score of 67\%, nearly matching the 67.2\% achieved by the \textbf{COT-ZeroShot}. A detailed analysis appears in Appendix~\ref{app:utilisation_of_mcg}.

\begin{figure}[t]
\centering
\includegraphics[width=0.7\linewidth]{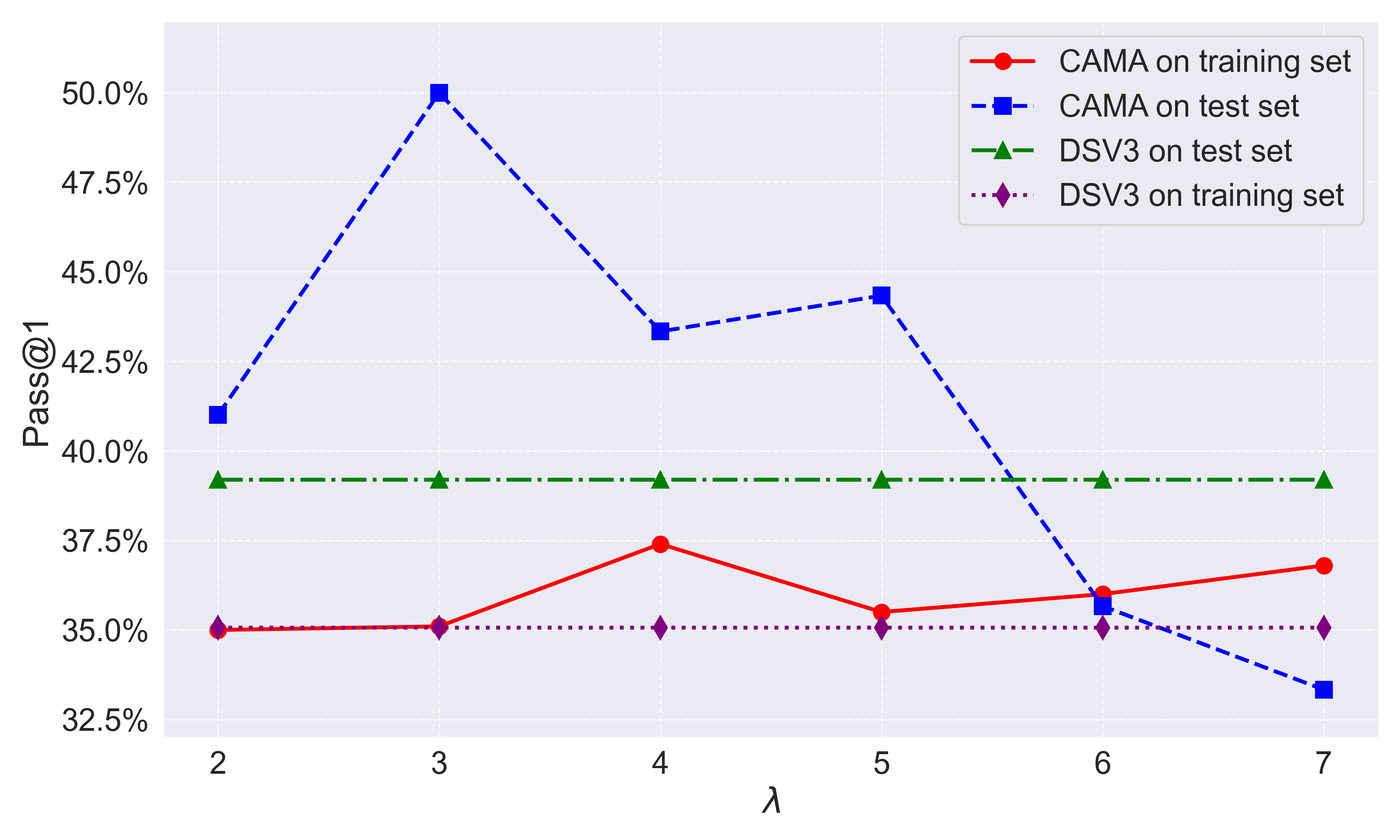} 
\caption{This figure shows the Pass@1 scores of \textbf{CAMA} using Mathematical Causal Graphs (MCGs) built with different knowledge point granularities, controlled by the parameter $\lambda$ (ranging from 2 to 7). Each $\lambda$ produces a distinct MCG from the AIME2022 and AIME2023 training data, and results are averaged over three repetitions. Performance is reported on both the training set (AIME2022 and AIME2023) and the test set (AIME2024). The base LLM is \textbf{DSV3}, with its scores included as references, shown with purple diamonds for the training sets and a green triangle for the test set.}
\label{fig:scaling_law_k}
\end{figure}

\paragraph{Impact of Knowledge Point Granularity $\lambda$}
In this part, we investigate how the granularity of knowledge points, controlled by $\lambda$, influences the performance of \textbf{CAMA}. For this experiment, we use AIME2022 and AIME2023 as the training set and AIME2024 as the test set. The value of $\lambda$ is varied from 2 to 7 in increments of 1. For each setting, an MCG is constructed from the training data and then reused to answer both training and test questions. Figure~\ref{fig:scaling_law_k} reports the Pass@1 scores of \textbf{CAMA} based on \textbf{DSV3} on both sets, with each result averaged over three repetitions. The results reveal an interesting finding: as $\lambda$ increases, performance improves on the training set but declines on the test set. This suggests a trade-off between generalization and specificity of the extracted knowledge points. When $\lambda$ is small, the extracted knowledge points are coarser and more broadly applicable across datasets. In contrast, larger $\lambda$ values lead to finer-grained, more detailed knowledge that better fits the training questions but may overfit, reducing robustness on unseen problems. Notably, the $\lambda$ that maximizes training performance is not necessarily optimal for the test set.

\paragraph{Case Study}
We apply \textbf{CAMA} with \textbf{DSV3} and focus on Problem 14 from Exam II of the AIME2024, which states:

\begin{quote}
    Let $b \geq 2$ be an integer. Call a positive integer $n$ $b$\textit{-eautiful} if it has exactly two digits when expressed in base $b$, and these two digits sum to $\sqrt{n}$. For example, $81$ is $13$-eautiful because $81=\underline{6}\underline{3}_{13}$ and $6+3=\sqrt{81}$. Find the least integer $b \geq 2$ for which there are more than ten $b$-eautiful integers.
\end{quote}

The reasoning trace produced by \textbf{DSV3} matches two knowledge points in the MCG: \textit{Modular arithmetic for integer solutions} and \textit{Quadratic polynomial systems}, with the former serving as a prerequisite for the latter. This dependency plays an important role in guiding \textbf{DSV3}'s problem-solving approach. \textbf{DSV3} first represents the two-digit number in base $b$ as $a_1\cdot b + a_0$, where $1 \leq a_1 \leq b-1$ and $0 \leq a_0 \leq b-1$, and imposes the constraint $(a_1 + a_0)^2 = a_1\cdot b + a_0$. Leveraging \textit{Modular arithmetic for integer solutions}, \textbf{DSV3} introduces $s = a_1 + a_0$, leading to the equation $a_1 = \frac{s(s - 1)}{b - 1}$. Since $s$ and $s - 1$ are consecutive integers and hence coprime, this implies that the denominator $b - 1$ should factor into a coprime pair $(d, e)$ such that $s \equiv 0 \pmod{d}$ and $s - 1 \equiv 0 \pmod{e}$. Subsequently, within the framework of \textit{Quadratic polynomial systems}, \textbf{DSV3} invokes the Chinese Remainder Theorem, and \textbf{DSV3} deduces that each coprime pair $(d, e)$ yields a distinct solution for $s$ that repeats every $b-1$ values. Since the number of such coprime factorizations $(d, e)$ is $2^{\gamma}$, where $\gamma$ is the number of distinct prime factors of $b - 1$, \textbf{DSV3} continues with additional derivations and ultimately returns the correct solution 211. The prompt is provided in Appendix~\ref{app:case_study}. However, without the incorporation of this prior knowledge, \textbf{DSV3} does not apply modular arithmetic and the Chinese Remainder Theorem, but instead resorts to an exhaustive brute-force search over possible digit pairs and ultimately returns an incorrect answer. Moreover, modular operations are generally essential for solving constrained quadratic polynomial functions, highlighting their broader relevance beyond this specific context.
\section{Conclusions}
In this paper, we present the Mathematical Causal Graph (MCG) as a high-level representation of reusable solution strategies for mathematical problems and introduce CAusal MAthematician (CAMA), a plug-and-play framework that learns and exploits these graphs. CAMA extracts knowledge points from chain-of-thought solutions generated by a large language model, infers the causal relations among them, and refines the resulting graph with feedback from its own question-answering accuracy. Experiments on real-world benchmarks show that incorporating structured guidance consistently outperforms unstructured alternatives, and that incorporating asymmetric causal relationships yields greater improvements than relying on symmetric associations alone. 

Although our proposed framework shows significant improvement, it still has some limitations. CAMA’s effectiveness depends on the granularity parameter $\lambda$, which controls how knowledge points are extracted. The optimal value of $\lambda$ on training data may not transfer well to unseen benchmarks, posing challenges for practical deployment. Future work could explore strategies for automatically selecting an appropriate $\lambda$ for new tasks, potentially with assistance from the LLM itself. Finally, the MCG remains static during the reasoning stage. Enabling CAMA to dynamically update the graph while answering questions, particularly by adding novel knowledge points not observed during training, could improve both its coverage and robustness.

\newpage
\bibliographystyle{unsrt}  
\bibliography{references}  


\clearpage
\onecolumn
\appendix

\setcounter{secnumdepth}{3}  

\renewcommand{\thesection}{\Alph{section}}
\renewcommand{\thesubsection}{\Alph{section}.\arabic{subsection}}

\section{Pseudocode for the CAMA Framework}
\label{app:pseudocode}
\subsection{Learning Stage of CAMA}
\label{app:code_learning}
\begin{algorithm}[!ht]
\caption{The learning stage of CAMA}
\label{alg:learning_CAMA}
\textbf{Input}: Dataset of question–solution–answer triples $\textbf{D} = \{(q_i, s_i, a_i)\}_{i=1}^n$; knowledge point granularity $\lambda$; a large language model $\mathcal{LLM}$;  
prompt templates: factor proposal $p_p$, redundancy removal $p_r$, answer generation $p_a$, task analysis $p_t$, subgraph matching $p_m$, and graph update $p_u$;  
causal discovery algorithm $\mathrm{CD}(\cdot)$;  
number of alignment QA samples $m$, batch size $s_b$, number of epochs $n_e$, history length $r$, and early stopping threshold $c_{\text{stop}}$. \\
\textbf{Output}: Optimized MCG $\mathcal{G}^*$

\begin{algorithmic}[1]
\Statex \textcolor{gray}{// \textbf{Initial MCG Construction}}
\FOR{each $(q_i, s_i)$ in $\textbf{D}$}
    \STATE $\mathbf{V}^i \gets \mathcal{LLM}(q_i, s_i, \lambda, p_p)$ \hfill \textit{// Extract up to $\lambda$ knowledge points}
\ENDFOR
\STATE $\mathbf{V}' \gets \bigcup_{i=1}^{n} \mathbf{V}^i$ \hfill \textit{// Aggregate raw knowledge points}
\STATE $(\mathbf{V}, \mathbf{O}) \gets \mathcal{LLM}(\mathbf{V}', p_r)$ \hfill \textit{// Deduplicate using LLM}
\STATE $\mathbf{Z} \gets l(\{\mathbf{V}^1, \dots, \mathbf{V}^n\}, \mathbf{V}, \mathbf{O})$ \hfill \textit{// Construct binary matrix}

\STATE $\mathcal{G}_1 \gets \mathrm{CD}(\mathbf{Z})$  \hfill \textit{// Mathematical Causal Graph Construction}

\Statex \textcolor{gray}{// \textbf{MCG Alignment via Question–Answer Feedback}}
\STATE Sample $\mathbf{D}^{s} = \{(q_i, s_i, a_i)\}_{i=1}^{m} \subset \mathbf{D}$
\STATE Initialize optimization history $\mathbf{H} \gets \emptyset$
\STATE Initialize round number $o \gets 1$
\STATE Initialize $\mathcal{G}^* \gets \mathcal{G}_1$, best precision $\mathcal{L}^* \gets 0$
\FOR{$\alpha = 1$ to $n_e$}
    \FOR{$\beta = 1$ to $\left\lceil \frac{m}{s_b} \right\rceil$}
        \STATE $o \gets (\alpha-1)\left\lceil \frac{m}{s_b} \right\rceil+\beta$
        \STATE Sample batch $\mathbf{D}'_o = \{(q_j, s_j, a_j)\}_{j=1}^{s_b} \subset \mathbf{D}^{s}$
        \FOR{each $(q_j, a_j)$ in $\mathbf{D}'_o$}
            \STATE $t_j \gets \mathcal{LLM}(q_j, p_t)$ \hfill \textit{// Generate reasoning trace}
            \STATE $\mathcal{G}^j_{o} \gets \mathcal{LLM}(\mathcal{G}_o, t_j, q_j, p_m)$ \hfill \textit{// Extract subgraph}
            \STATE $\hat{a}_j \gets \mathcal{LLM}(\mathcal{G}^j_{o}, q_j, p_a)$ \hfill \textit{// Predict answer}
        \ENDFOR
        \STATE $\mathcal{G}_{o+1} \gets \mathcal{LLM}(\{(q_j, s_j, \mathcal{G}^j_{o}, \mathbf{1}_{\{\hat{a}_j = a_j\}})\}_{j=i}^{s_b}, \mathbf{H}, p_p)$ \hfill \textit{// Graph update}
        \STATE $\mathcal{L}_o \gets \frac{1}{s_b} \sum_{j=1}^{s_b} \mathbf{1}_{\{\hat{a}_j = a_j\}}$ \hfill \textit{// Compute batch precision}
        \STATE $\mathbf{H} \gets \mathbf{H} \cup \{(\mathcal{G}_{o}, \mathcal{L}_{o})\}$ \hfill \textit{// Update optimization history}
        \IF{No change in $\mathcal{G}_{o+1}$ for $c_{\text{stop}}$ consecutive rounds}
            \STATE \textbf{break}
        \ENDIF
    \ENDFOR
    \STATE $\mathcal{L}_{\text{full}} \gets \frac{1}{m} \sum_{i=1}^{m} \mathbf{1}_{\{\mathcal{LLM}(\mathcal{G}_{o}, q_i, p_a) = a_i\}}$ \hfill \textit{//Evaluate full precision $\mathcal{L}_{\text{full}}$ of $\mathcal{G}_{o}$ on $\mathbf{D}^s$}
    \IF{$\mathcal{L}_{\text{full}} > \mathcal{L}^*$}
        \STATE $\mathcal{G}^* \gets \mathcal{G}_{o}$
        \STATE $\mathcal{L}^* \gets \mathcal{L}_{\text{full}}$
    \ENDIF
\ENDFOR

\STATE \textbf{return} $\mathcal{G}^*$
\end{algorithmic}
\end{algorithm}

\subsection{Reasoning Stage of CAMA}
\label{app:code_reasoning}
\begin{algorithm}[H]
\caption{The reasoning stage of CAMA}
\label{alg:reasoning_CAMA}
\textbf{Input}: Question $q_j$; MCG $\mathcal{G}^*$; a large language model $\mathcal{LLM}$;  
prompt templates: task analysis $p_t$, subgraph matching $p_m$, answer generation $p_a$\\
\textbf{Output}: Predicted answer $\hat{a}_j$
\begin{algorithmic}[1]
\Statex \textcolor{gray}{// \textbf{Generate Reasoning Trace}}
\STATE $t_j \gets \mathcal{LLM}(q_j, p_t)$
\Statex \textcolor{gray}{// \textbf{Extract Relevant Subgraph}}
\STATE $\mathcal{G}^j \gets \mathcal{LLM}(\mathcal{G}^*, t_j, q_j, p_m)$ 
\Statex \textcolor{gray}{// \textbf{Answer the Question}}
\STATE $\hat{a}_j \gets \mathcal{LLM}(\mathcal{G}^j, q_j, p_a)$

\STATE \textbf{return} $\hat{a}_j$
\end{algorithmic}
\end{algorithm}

\section{The Learned Mathematical Causal Graph}
\label{app:mcg}
\begin{figure}[!ht]
\centering
\includegraphics[width=0.9\columnwidth]{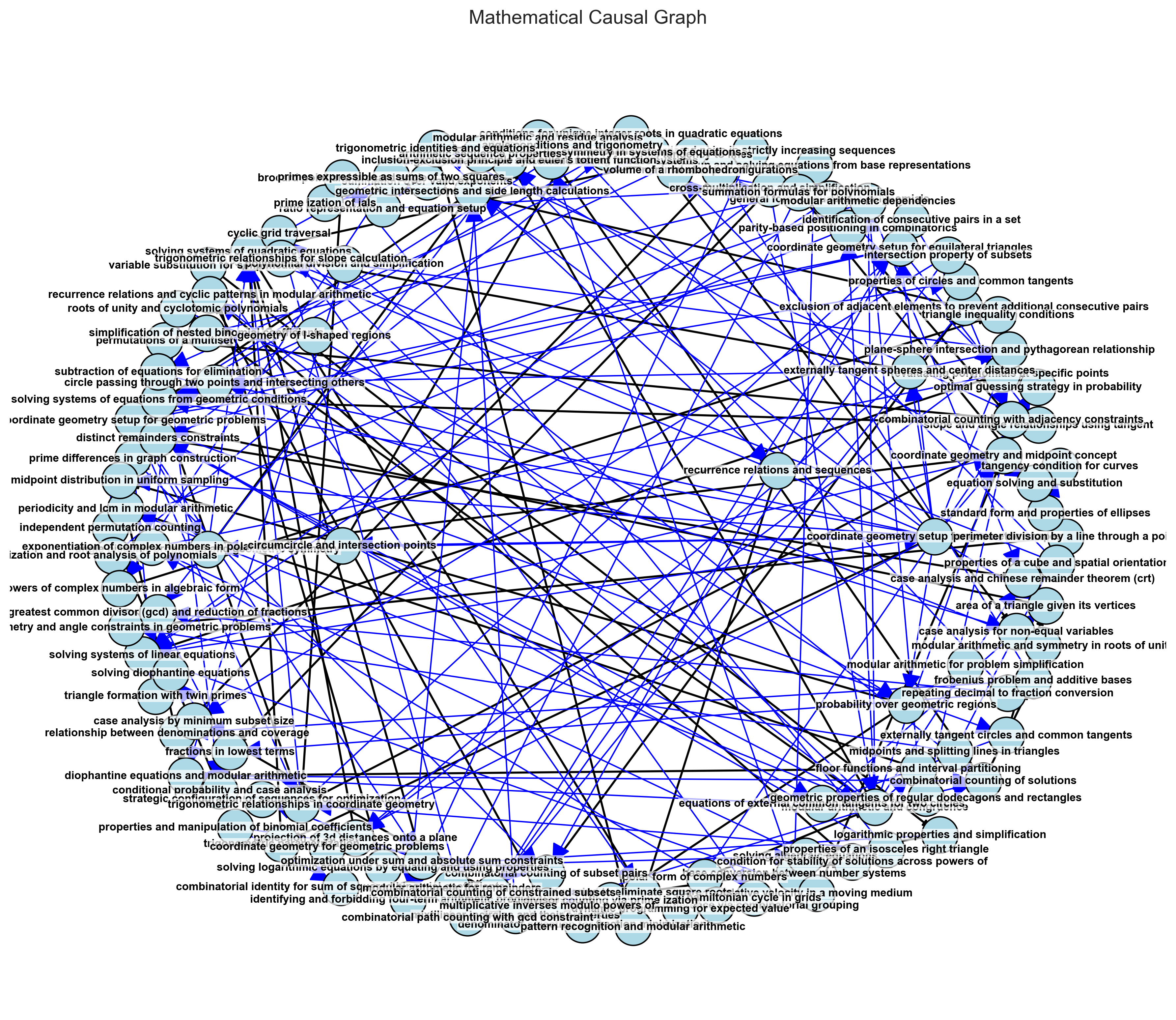} 
\caption{When the knowledge point granularity is set to $\lambda=3$, the Mathematical Causal Graph learned by CAMA from the combined AIME2022 and AIME2023 datasets consists of 124 nodes and 184 edges, including 129 directed edges (in blue) and 55 undirected edges (in black).}
\label{fig:mcg}
\end{figure}

Figure \ref{fig:mcg} presents the Mathematical Causal Graph learned by CAMA on DSV3 from the combined AIME2022 and AIME2023 datasets with knowledge point granularity $\lambda=3$. The graph comprises 124 nodes and 184 edges, including 129 directed edges (in blue) and 55 undirected edges (in black).

\section{Additional Results}

\subsection{Standard Deviation of Pass@1 Scores Across Four Datasets}
\label{app:std_passe_1_score}
Table~\ref{tab:std_pass1} reports the standard deviation of Pass@1 scores for each method on four datasets using DeepSeek-V3-0324 (\textbf{DSV3}) and Qwen3-32B (\textbf{Qwen3}). Each value is averaged over three repetitions. Overall, across both base LLMs, \textbf{CAMA} exhibits stable and comparable variability relative to other baselines and its ablated variants.

\begin{table*}[!h]
\centering
\begin{tabular}{c|c|c|c|c|c}
\toprule
\multirow{2}{*}{Base LLM} & \multirow{2}{*}{Method} &  AIME2024 & AIME2025 & Omni-MATH-200 & OlympiadBench-674 \\ 
&  & Pass@1($\uparrow$) & Pass@1($\uparrow$) & Pass@1($\uparrow$) & Pass@1($\uparrow$)  \\ \midrule  
\multirow{6}{*}{\textbf{DSV3}} & \textbf{COT-ZeroShot} & 3.1\% & 3.1\% & 1.4\% & 2.3\% \\
& \textbf{COT-FewShot} & 1.6\% & 3.2\% & 1.8\% & 0.3\% \\
& \textbf{DSV3} & 1.6\% & 4.2\% & 0.8\% & 0.8\% \\
& \textbf{CAMA \textit{w/o} Directed Edge} & 2.7\% & 5.4\% & 1.2\% & 0.6\% \\
& \textbf{CAMA \textit{w/o} Alignment} & 3.1\% & 4.2\% & 1.5\% & 1.3\% \\
& \textbf{CAMA} (ours) & 2.7\% & 1.6\% & 1.9\% & 0.6\%\\
\midrule
\multirow{6}{*}{\textbf{Qwen3}} & \textbf{COT-ZeroShot} & 4.2\% & 3.1\% & 1.5\% & 0.7\% \\
& \textbf{COT-FewShot} & 2.7\% & 4.2\% & 2.5\% & 0.1\%\\
& \textbf{Qwen3} & 1.6\% & 5.7\% & 1.8\% & 0.5\% \\
& \textbf{CAMA \textit{w/o} Directed Edge} & 6.8\% & 1.6\% & 1.1\% & 0.7\% \\
& \textbf{CAMA \textit{w/o} Alignment} & 1.6\% & 1.6\% & 2.9\% & 0.7\% \\
& \textbf{CAMA} (ours) & 1.6\% & 2.7\% & 0.7\% & 0.4\% \\
\bottomrule
\end{tabular}

\caption{This table reports the standard deviation of Pass@1 scores for each method on four datasets using DeepSeek-V3-0324 (\textbf{DSV3}) and Qwen3-32B (\textbf{Qwen3}). Each value is averaged over three repetitions. Omni-MATH-200 is a 200-question subset of Omni-MATH covering four categories. Likewise, OlympiadBench-674 denotes a 674-question English, non-proof subset of OlympiadBench. Bold values indicate better performance.}
\label{tab:std_pass1}
\end{table*}

\subsection{Question Distribution Across Four Mathematical Datasets}
\label{app:dataset_distribution}
Table~\ref{tab:fine_grained_analyze} presents the distribution of questions in the four mathematical datasets AIME2024, AIME2025, Omni-MATH-200, and OlympiadBench-674 across the categories algebra (\textbf{A}), geometry (\textbf{G}), number theory (\textbf{NT}), and combinatorics (\textbf{C}).

\begin{table}[!ht]  
\centering
\begin{tabular}{c|cccc}
 & \textbf{A} & \textbf{G} & \textbf{NT} &  \textbf{C}  \\
 \midrule
AIME2024 & 26.7\% & 23.3\% & 26.7\% & 23.3\% \\
AIME2025 & 26.7\% & 23.3\% & 26.7\% & 23.3\% \\
Omni-MATH-200 & 25.0\% & 25.0\% & 25.0\% & 25.0\% \\
OlympiadBench-674 & 39.2\% & 19.2\% & 18.8\% & 22.8\% \\
\end{tabular} 
\caption{Percentage distribution of questions in four mathematical datasets across the categories algebra (\textbf{A}), geometry (\textbf{G}), number theory (\textbf{NT}), and combinatorics (\textbf{C}).}
\label{tab:fine_grained_analyze}
\end{table}

\subsection{MCG Coverage Across Test Sets}
\label{app:utilisation_of_mcg}

Table~\ref{tab:num_matched_KP} reports the extent to which the Mathematical Causal Graph (MCG) learned by CAMA contributes to answering questions across different test sets when the granularity parameter is fixed at $\lambda=3$. For OlympiadBench-674, the MCG supports only 3\% of the questions, with an average of 1 matched knowledge point per supported question. In contrast, for AIME2024, AIME2025, and Omni-MATH-200, the MCG is utilized in 100\% of the questions, with an average of at least 4.3 matched knowledge points per question. These results help explain CAMA’s comparatively moderate performance on OlympiadBench-674: the learned MCG lacks sufficient coverage of the knowledge components required by OlympiadBench-674. 

\begin{figure*}[!ht]
  \centering
  \begin{minipage}[t]{0.48\textwidth}
    \centering
    \includegraphics[width=\linewidth]{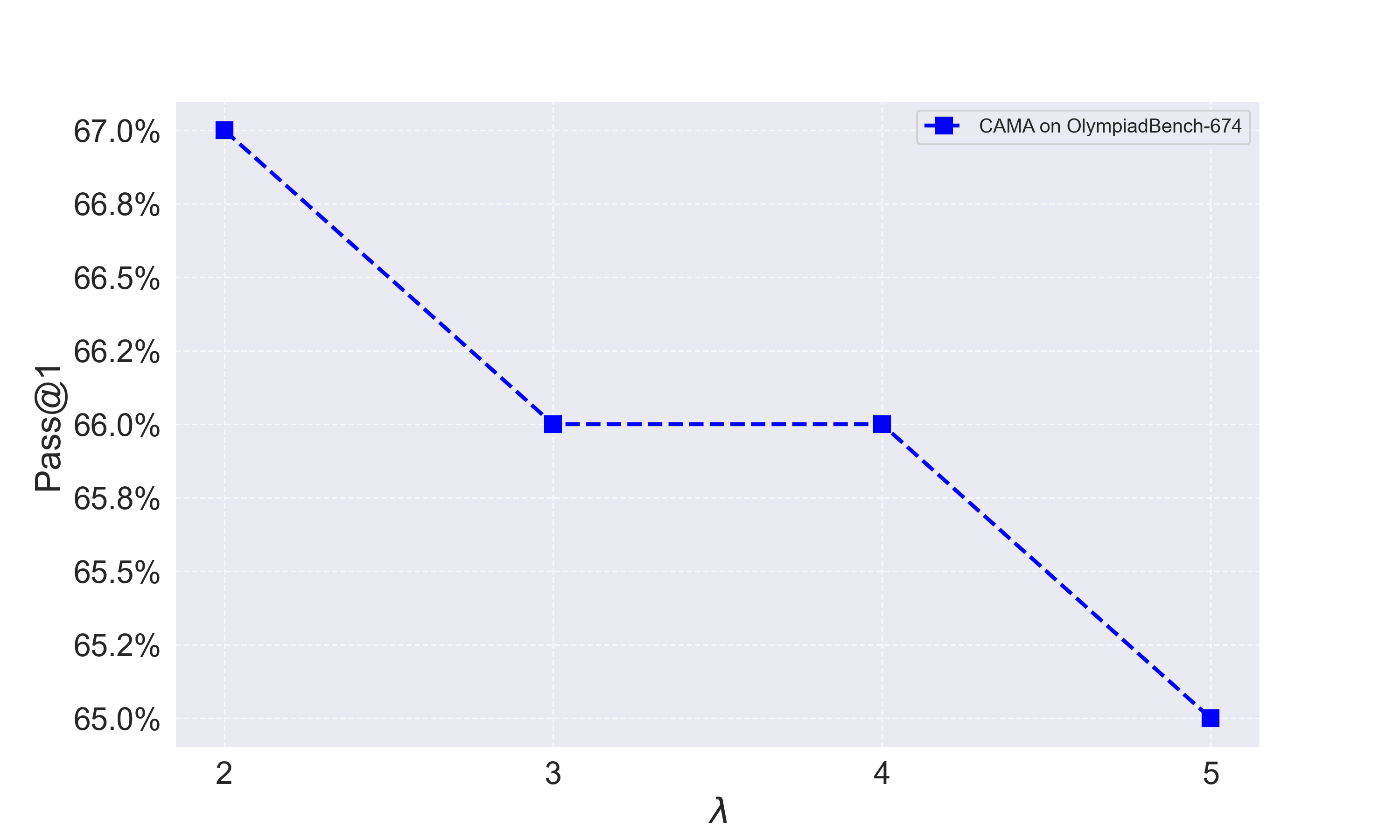}
  \end{minipage}
  \begin{minipage}[t]{0.48\textwidth}
    \centering
    \includegraphics[width=\linewidth]{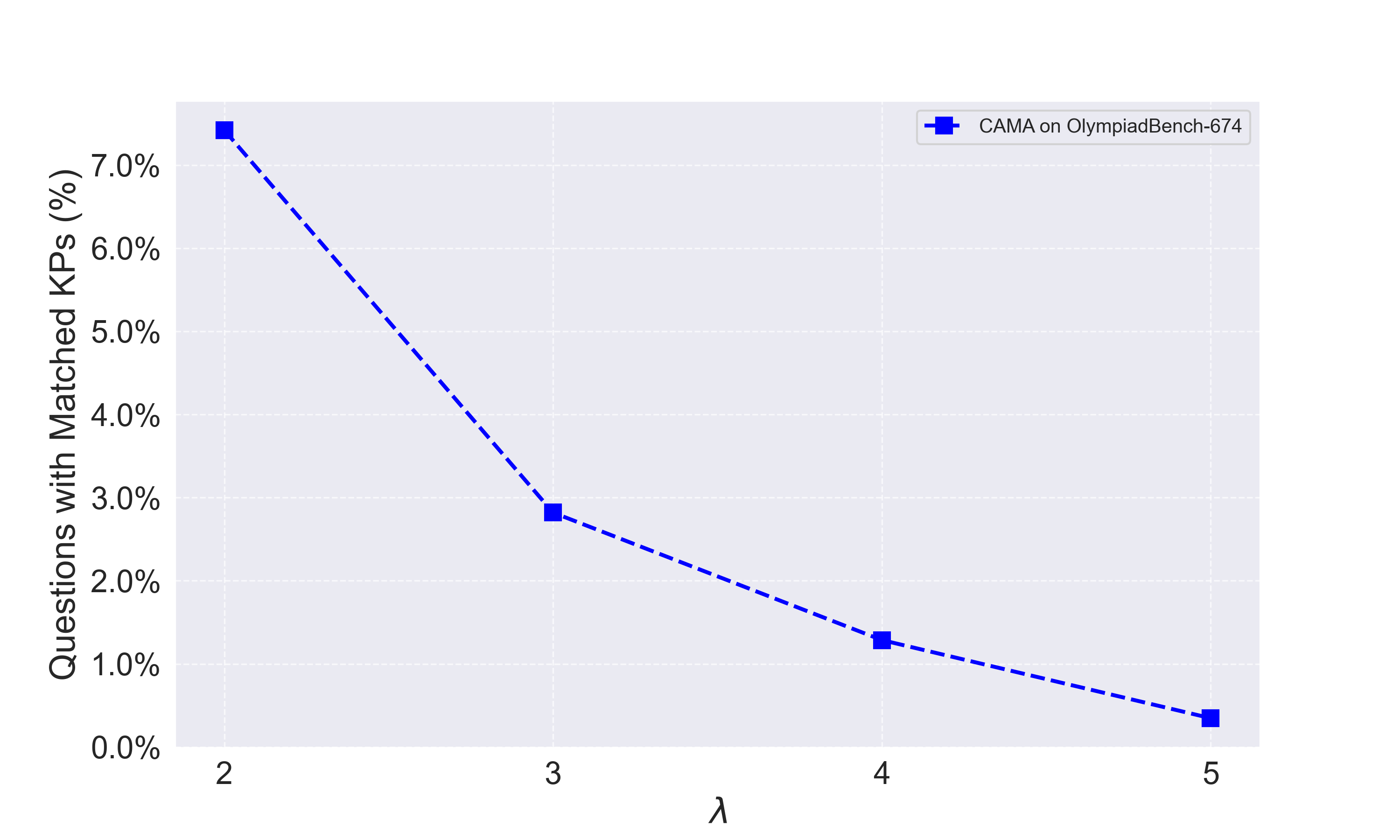} 
  \end{minipage}
  \caption{Left: Pass@1 score of CAMA on OlympiadBench-674 with MCGs built at different granularities ($\lambda=2\text{--}5$). Right: percentage of OlympiadBench-674 questions that use at least one knowledge point present in each MCG. Results are averaged over three repetitions.}
  \label{fig:OB_lambda}
\end{figure*}

Because $\lambda$ controls the granularity of the extracted knowledge points, smaller values yield more general concepts that transfer to unseen datasets, whereas larger values produce finer, dataset‑specific points. We therefore built MCGs with $\lambda$ ranging from 2 to 5, trained on AIME2022 and AIME2023, and evaluated them on OlympiadBench‑674.  The left panel of Figure~\ref{fig:OB_lambda} plots Pass@1 score, and the right panel shows the percentage of questions with at least one matched knowledge point. As $\lambda$ increases, both Pass@1 and the knowledge point utilization decline. When $\lambda=2$, CAMA attains a Pass@1 score of 67.0\%, virtually matching the 67.2\% of the CoT‑ZeroShot, while also delivering the highest knowledge point utilization. This result suggests that the MCG built at this granularity better captures the knowledge required for OlympiadBench‑674 than the graph obtained with $\lambda = 3$. 

\begin{table*}[!ht]
\centering
\begin{tabular}{c|c|c|c|c}
  &  AIME2024 & AIME2025 & Omni-MATH-200 & OlympiadBench-674 \\
  \midrule
 Questions with Matched KPs (\%) & 100\% & 100\% & 100\% & 3\%  \\
 Matched KPs per Question (with Match) & 4.3 & 4.6 & 5.5 & 1   \\
\end{tabular}
\caption{Utilization of the Mathematical Causal Graph (MCG) across test sets. Metrics include: (1) the percentage of questions with at least one matched knowledge point (KP) and (2) the average number of matched KPs per question among those with at least one matched KP. Each value is averaged over three repetitions. }
\label{tab:num_matched_KP}
\end{table*}

\section{Prompt Used in the Case Study}
\label{app:case_study}
The following prompt was employed in the case study to address Problem 14 from Exam II of the AIME 2024: 

\begin{lstlisting}[language=] 
# Question 

Let $b \geq 2$ be an integer. Call a positive integer $n$ $b$\textit{-eautiful} if it has exactly two digits when expressed in base $b$, and these two digits sum to $\sqrt{n}$. For example, $81$ is $13$-eautiful because $81=\underline{6}\underline{3}_{13}$ and $6+3=\sqrt{81}$. Find the least integer $b \geq 2$ for which there are more than ten $b$-eautiful integers.

# Elements to Consider:

 **1.** knowledge point 16 modular arithmetic for integer solutions \\Using congruence relations (e.g., solving $(ax \equiv b \pmod{n})$) and Diophantine equations to find valid integer solutions under constraints.
 
 **2.** knowledge point 1 quadratic polynomial systems \\Understanding how to set up and solve systems of equations derived from given points and leading coefficients of quadratic polynomials. This includes leveraging properties like vertex form, leading coefficients, and substitution methods.

# Relationship(s) Among Element(s):

 **1.** knowledge point 16 modular arithmetic for integer solutions directly influences or is a prerequisite for knowledge point 1 quadratic polynomial systems. If knowledge point 1 quadratic polynomial systems used, then knowledge point 16 modular arithmetic for integer solutions should also be used.

# Task

Consider the question carefully and work through the solution step by step, keeping in mind the elements provided and any relationships between them.
For the output, your reasoning should be enclosed in <think> </think> tags, and the final answer should be enclosed in <answer> </answer> tags.

The final answer must strictly follow the format:

**"The answer is: ___."**
\end{lstlisting}


\section{Details of Additional Prompts Used in the Paper}
\label{app:used_prompts}
This section provides the prompt templates employed in various components of the CAMA framework.
\subsection{Mathematical Dataset Construction}
The prompt used to generate a detailed chain-of-thought solution $s_i$ and a predicted answer $\hat{a}_i$
is denoted by $p_g$. It is parameterized by a single input, \texttt{question}, which represents the raw text of the mathematical question.

\begin{lstlisting}[language=]
# Question 

{question}

# Task

Carefully analyze the question and break it down step by step. 

**For the output, your reasoning should be enclosed in <think> </think> tags, and the final answer should be enclosed in <answer> </answer> tags.**

\end{lstlisting}

\subsection{Learning Stage}

\paragraph{Initial MCG Construction}
The prompt used to extract up to $\lambda$ knowledge points from each question is denoted by $p_p$. It is parameterized by two inputs: \texttt{question\_solution\_pairs}, which represents a set of mathematical question–solution pairs, and \texttt{lambda}, which specifies the desired granularity of the extracted knowledge points.

\begin{lstlisting}[language=]  
# What is a Knowledge Point?
A knowledge point is a self-contained mathematical concept, technique, or principle that is applied in solving the problem. Below are two examples:

1. pythagorean theorem:\n The pythagorean theorem states that in a right-angled triangle, the square of the length of the hypotenuse (the side opposite the right angle) is equal to the sum of the squares of the lengths of the other two sides. Mathematically, if $a$ and $b$ are the legs of a right triangle and $c$ is the hypotenuse, then the relationship is given by: $a^2 + b^2 = c^2$.
2. polar form of complex numbers:\n The polar form of a complex number expresses the number in terms of its magnitude and angle (also called modulus and argument). If a complex number is given as $z = x + iy$, where $x$ is the real part and $y$ is the imaginary part, then it can also be represented as:$z = r (\cos \\theta + i \sin \\theta)$, where $r = |z| = \sqrt{{x^2 + y^2}}$ is the modulus of $z$, and $\\theta = \\arg(z) = \\tan^{{-1}}\left(\\frac{{y}}{{x}}\\right)$ is the argument (angle) of $z$. This form is also commonly written using Euler's formula as: $z = r e^\\{{i\\theta\\}}$.

# Input
A math question and its correct solution are provided below:

{question_solution_pairs}

# Task: Extract Key Knowledge Points
Your task is to **analyze the question and its solution** to extract **up to {lambda} distinct and essential knowledge points** required to solve the problem correctly. Refer to the two examples above as a guide for the level of detail and clarity expected.

Follow these steps:

1. **Identify up to {lambda} general, relevant knowledge points** that play a key role in solving the problem.
2. Each knowledge point should represent a **unique concept, skill, or method** used in the solution.
3. **Avoid redundancy**-each point should address a different aspect of the problem-solving process.

# About Output

Your output should include the following parts:

**Part 1**: Reasoning Process.

Describe your thought process for identifying and designing knowledge points. Consider the following: 
- Extract key ideas from the questions and solutions to form appropriate knowledge points.
- Clearly define the criteria for each knowledge point, explaining why it is relevant.

**Part 2**: Knowledge Points Filtration. 

You shoud decide whether to use each of the proposed knowledge points by following criteria:
- The knowledge point should contribute to correctly answering the question.
- Each knowledge point should focus on a specific aspect, avoiding overlap with other points.

**Part 3**: Final Output. 

Report the final list of knowledge points you have selected.
- **For each knowledge point, assign a clear and concise name, and provide a detailed description of its role-without referencing any specific question index.**

Report the factors **in following template:**

```
**Knowledge Point Name**: [Description of this knowledge point].
```
\end{lstlisting}

The prompt used to remove redundancy from the aggregated set of raw knowledge points is denoted by $p_r$. It is parameterized by one variable: \texttt{list\_all\_knowledge\_points}, which represents the full set of extracted knowledge points prior to deduplication.

\begin{lstlisting}[language=]
# Input:

You are given a list of knowledge points along with their descriptions:

{list_all_knowledge_points}

# Task:

Some knowledge points in the list may be redundant - meaning they describe the same or very similar concepts. Your task is to:

1. Carefully analyze the list to identify any redundant knowledge points.
2. Determine which of the remaining knowledge points can **replace** the redundant ones.
3. Ensure that:
   * Each knowledge point marked as a replacement is **not** included in the removed list.
   * A single knowledge point may replace multiple redundant ones.

# Output Format:

* Wrap your reasoning in a <think>...</think> block.
* Present your final answer in an <answer>...</answer> block using the format below.
* Only use the **names** of the knowledge points (not their descriptions) in the output.

```
<answer>
**Removed Knowledge Points:**
[**<Knowledge Point A>**, **<Knowledge Point B>**, ...]

**Replacement Details:**
[**<Knowledge Point C>** can replace **<Knowledge Point A>**,
 **<Knowledge Point D>** can replace **<Knowledge Point B>**,
 ...]
</answer>
```
\end{lstlisting}

\paragraph{MCG Alignment via Question–Answer Feedback}
The prompt used to update the Mathematical Concept Graph (MCG) based on question–answer feedback within each batch is denoted by $p_u$. It is parameterized by two inputs: \texttt{qa\_correct\_answer}, which includes the original questions, correct solutions, the matched knowledge points, and their inferred causal dependencies for correctly answered instances; and \texttt{qa\_incorrect\_answer}, which provides the same information for instances that were answered incorrectly.

\begin{lstlisting}[language=]
# Input Data

The following two sets of question-answer pairs are provided:

* **Correctly Answered Questions**
  Each entry includes the original question, the correct solution, previously matched knowledge points, and currently recorded relations between those knowledge points (if any).
  `{qa_correct_answer}`

* **Incorrectly Answered Questions**
  Each entry includes the original question, the correct solution, previously matched knowledge points, and currently recorded relations between those knowledge points (if any).
  `{qa_incorrect_answer}`

# Task Instructions

Review the provided examples and perform the following steps:

1. **Analyze each question and its solution**, carefully considering how the different knowledge points are applied in solving the problem.
2. **Revise or correct the relationships between the knowledge points** listed in each example.
3. For each pair that **requires modification**, specify the corrected relationship using one of the following categories:

   * **Prerequisite** - Knowledge Point A must be used or understood before Knowledge Point B.
   * **Dependent** - The two knowledge points are conceptually or procedurally linked, but there's no clear ordering.
   * **Independent** - The two knowledge points are unrelated or not used together in this problem.

You may find some existing relations are incorrect or too vague; refine them for accuracy and utility.

---

# Output Format

Wrap your thinking process inside <think>...</think> and present your final answer inside <answer>...</answer>, formatted as a list of statements-one per line-using the structure below:

```
[**<Knowledge Point A>** is prerequisite/dependent/independent of **<Knowledge Point B>**.
 **<Knowledge Point C>** is prerequisite/dependent/independent of **<Knowledge Point D>**.
 ...]
```

Use **"prerequisite"** when Knowledge Point A must be understood before using Knowledge Point B in the context of solving the problem.
Use **"dependent"** if A and B are used together or closely related in solving the problem, but there is no clear applying order between them.
Use **"independent"** if A and B are not necessarily related or used together in the problem-solving context.

---

**Note:**

* Focus only on relationships that **need correction or clarification**.
* Avoid repeating vague or incorrect relations from the input - your role is to improve precision.
* Ensure that your suggested relations would help a learner understand **how to approach and solve similar problems** more effectively.

\end{lstlisting}

\subsection{Reasoning Stage}
\paragraph{Generate Reasoning Trace}
The prompt used to analyze a question and generate a corresponding reasoning trace is denoted by $p_t$. It is parameterized by a single input, \texttt{question}, which represents the raw text of the mathematical question.

\begin{lstlisting}[language=]
# Question 

{question}

# Task

Carefully analyze the question and break it down step by step to identify the key concepts or elements required to solve the problem. 

**For the output, your reasoning should be enclosed in <think> </think> tags, and the final answer should be enclosed in <answer> </answer> tags.**
\end{lstlisting}

\paragraph{Extract Relevant Subgraph}
The prompt used to match relevant knowledge points for answering a given question is denoted by $p_m$. It is parameterized by two inputs: \texttt{question\_think}, which includes the mathematical question along with its associated reasoning trace, and \texttt{knowledge\_point\_descriptions}, which provides the list of candidate knowledge points and their corresponding descriptions.
\begin{lstlisting}[language=]
# Problem

{question_think}

# List of Factors 

{knowledge_point_descriptions}

# Task

Carefully read the problem and the reasoning behind it. Then, select the relevant factors from the list above that could help solve the problem.

# Response Format
Your response should be in the following format, listing the indices of the chosen factors:

**The chosen factors are: [Index of factor 1, Index of factor 2, ...].**

\end{lstlisting}

\paragraph{Answer the Question}
The prompt used for re-answering the question is denoted by $p_a$. It is parameterized by three inputs: \texttt{question}, which represents the input mathematical question; \texttt{chosen\_knowledge\_points}, which specifies the selected knowledge points along with their corresponding descriptions; and \texttt{knowledge\_point\_relations}, which encodes the relationships among these knowledge points.

\begin{lstlisting}[language=]
# Question: 

{question}

# Elements to Consider:

{chosen_knowledge_points}

# Relationship(s) Among Element(s):

{knowledge_point_relations}

# Task

Consider the question carefully and work through the solution step by step, keeping in mind the elements provided and any relationships between them.
For the output, your reasoning should be enclosed in <think> </think> tags, and the final answer should be enclosed in <answer> </answer> tags.

The final answer must strictly follow the format:
**"The answer is: ___."**
\end{lstlisting}

\end{document}